\documentclass[conference]{IEEEtran}
\IEEEoverridecommandlockouts
\usepackage{cite}
\usepackage{amsmath,amssymb,amsfonts}
\usepackage{algorithmic}
\usepackage{graphicx}
\usepackage{textcomp}
\usepackage{tabularx}
\usepackage{xcolor}
\def\BibTeX{{\rm B\kern-.05em{\sc i\kern-.025em b}\kern-.08em
    T\kern-.1667em\lower.7ex\hbox{E}\kern-.125emX}}

\begin{document}
\bstctlcite{IEEEexample:BSTcontrol}

\title{Data Augmentation for Neural NLP}

\author{
Domagoj Plu{\v{s}}{\v{c}}ec \quad Jan {\v{S}}najder\\
University of Zagreb, Faculty of Electrical Engineering and Computing\\
Text Analysis and Knowledge Engineering Lab\\
Unska 3, 10000 Zagreb, Croatia \\
\tt \{domagoj.pluscec;jan.snajder\}@fer.hr
%\IEEEauthorblockN{Domagoj Pluščec}
%\and
%\IEEEauthorblockN{Jan Šnajder}
%\IEEEauthorblockA{\textit{Text Analysis and Knowledge Engineering Lab} \\
%\textit{Faculty of Electrical Engineering and Computing, University of Zagreb}\\
%Zagreb, Croatia \\
%domagoj.pluscec@fer.hr}
}

\maketitle

\begin{abstract}
Data scarcity is a problem that occurs in languages and tasks where we do not have large amounts of labeled data but want to use state-of-the-art models. Such models are often deep learning models that require a significant amount of data to train.
Acquiring data for various machine learning problems is accompanied by high labeling costs. Data augmentation is a low-cost approach for tackling data scarcity. 
This paper gives an overview of current state-of-the-art data augmentation methods used for natural language processing, with an emphasis on methods for neural and transformer-based models.
Furthermore, it discusses the practical challenges of data augmentation, possible mitigations, and directions for future research.
\end{abstract}

\begin{IEEEkeywords}
data augmentation, natural language processing, deep learning, low data regimes
\end{IEEEkeywords}

\section{Introduction}
Deep learning and transformer-based models have demonstrated excellent performance across many tasks in the field of natural language processing (NLP), such as machine translation \cite{Tan2020}, question answering \cite{Devlin2019}, text classification \cite{Minaee2021}, and language inference \cite{Devlin2019}. However, many of these models are trained using supervised learning, which requires a large amount of labeled data. Data labeling may be expensive and time-consuming. This problem is even more pronounced when such models are applied to tasks in low-resource languages \cite{Sahin2022}. 

One approach to mitigating data scarcity is data augmentation. Data augmentation is defined by \cite{Feng2021}, \cite{Li2022}, and \cite{Bayer2021} as a set of strategies for generating new data based on available data. Augmentation strategies can range from simple paraphrasing techniques that use linguistic resources, such as synonym replacement, to more complex sampling methods that try to generate new data based on the distribution of the original data.

Data augmentation strategies can help us solve several problems if applied correctly. Firstly, they can help improve the model's performance. Generated data can alleviate data scarcity in low-data regimes \cite{Hedderich2021}, add data diversity by injecting prior knowledge without changing the model \cite{Shorten2019}, and balance an imbalanced dataset by adding examples with not frequently occurring classes \cite{Bayer2021}. Some studies have found that data augmentation is a preferred regularization method \cite{Garcia2018} as it improves generalization without degrading the model's representational capacity or retuning other hyperparameters. Secondly, data augmentation strategies can help make the model more robust by securing it against adversarial examples \cite{Cheung2021} and simulating potential distribution shifts \cite{Shorten2021}. Transformer-based models tend to suffer from out-of-domain problem \cite{Sahin2022}, which can be alleviated by simulating domain shifts with data augmentation. Furthermore, data augmentation can help mitigate model bias \cite{Sharma2020}, \cite{Park2018} and substitute real-world data to remove personally identifiable information and protect peoples' privacy.

While data augmentation methods can be model agnostic, we will focus on the ones that work well with neural NLP models, with an emphasis on transformer-based models. Neural NLP models typically work on sequences of linguistic units that are represented as vectors. Compared to the more traditional models, neural NLP models do not require hand-crafted features and need large amounts of data to train. 
Some methods that were used traditionally in NLP, such as Easy Data Augmentation (EDA) \cite{Wei2019} and An Easier Data Augmentation (AEDA) \cite{Karimi2021}, showed poor performance when applied with transformer-based models. One possible explanation for the poor performance is that such pre-trained models saw a huge variety of data while pre-training \cite{Karimi2021}. Therefore, using data augmentation to add rarely occurring words and synonyms does not tend to improve model performance \cite{Karimi2021}. Longpre et al. \cite{Longpre2020} demonstrate that some data augmentation methods cannot achieve gains when using large pre-trained language models as they are already invariant to various transformations, such as synonym replacement. The authors also state that large language models map data to a latent space with representations nearly invariant to some transformations. As some of the traditional data augmentation methods do not work for neural models, it makes sense to investigate data augmentation for neural NLP specifically.

The remainder of the paper is structured as follows. Section \ref{Label preservation} explains the challenges of label preservation in data augmentation. Section \ref{Data augmentation methods} provides an overview of data augmentation methods with a focus on strategies suitable for deep learning models. Section \ref{Practical challenges} addresses the practical challenges of applying data augmentation methods.

\section{Label preservation}
\label{Label preservation}
In order for augmented data to be useful, it needs to be labeled correctly. We usually want to change a sample to add some diversity while also preserving the original labels. Alternatively, if augmentation changes the labels, we need to know how exactly the labels were changed.

It is easy to define label-preserving transformations for some data modalities. For instance, we can apply image rotation \cite{Simard2003} in computer vision or change the speed of sound \cite{Ko2015} in speech processing. However, changing data meaningfully is very challenging in NLP as text is discrete. Changing one word in a sentence could lead to changing the meaning of the entire sentence \cite{LiuP2020}. 

Many transformations come with no guarantees to preserve the correct label, but there is a high probability that the label will not change. This probability is referred to as the \emph{safety} of the data augmentation method \cite{Shorten2019}. Safer data augmentation methods do not necessarily achieve better performance. For example, random swap and random deletion, which are not label preserving, improve the model performance more than synonym replacement and insertion, which tend to be label preserving \cite{Bayer2021}.

A number of methods have been proposed to mitigate the risk of changing the sample's label. One approach is to select samples using a heuristic. For example, Alzantot et al. \cite{Alzantot2018} use a model that is trained on the original dataset to determine the confidence of the augmented sample. Only the samples that maximize the model's confidence are selected. While this approach does not improve the model performance because only the augmented samples that the original model predicts with maximum confidence are selected, it makes the model more robust against adversarial attacks.

Models that generate data can be conditioned to generate a sample for a particular label or determine the correct label for a newly generated sample. The c-BERT model \cite{Wu2019} uses BERT \cite{Devlin2019} segment embeddings to encode label information. When fine-tuned, the c-BERT model can be used to substitute a word from the original sample so that the resulting sample matches the sentence label. DataBoost \cite{Liu2020} is an example of applying reinforcement learning to generative models with the aim of generating a sentence that matches the selected class.

\section{Data augmentation methods}
\label{Data augmentation methods}
 Previous surveys proposed various categorizations of data augmentation methods. In this section, we will explore the different categorization. Firstly, we will discuss the categorization of data augmentation methods based on supervision (Section \ref{Supervision of data augmentation methods}. Secondly, we will discuss the categorization based on how the data augmentation is applied to the data (Section \ref{Data representation}). Finally, we will focus on taxonomy from \cite{Li2022}. The chosen taxonomy is based on the diversity of augmented samples and divides data augmentation methods into paraphrasing (Section \ref{Paraphrasing}), noising (Section \ref{Noising}), and sampling (Section \ref{Sampling}). 

Table \ref{tab:DataAugmentationMethods} shows examples of transformed data by using data augmentation methods described in this section.

\begin{table*}[h!t]
    \caption{Examples of transformed data by using data augmentation. Bold typeface indicates words that were transformed by data augmentation.}
   
    \begin{center}
\begin{tabularx}{\linewidth}{p{0.22\linewidth} X X}
\hline
Method &
  Original data&
  Augmented data
   \\
   \hline
 Generating adversarial examples for sentiment analysis task \cite{Alzantot2018} &
 This movie had \textbf{terrible} acting, \textbf{terrible} plot, and \textbf{terrible} choice of actors. (Leslie Nielsen ...come on!!!) the one part I \textbf{considered} slightly funny was the battling FBI/CIA agents, but because the audience was mainly \textbf{kids} they didn't understand that theme. (Negative, 78.0\%) &
This movie had \textbf{horrific} acting, \textbf{horrific} plot, and \textbf{horrifying} choice of actors. (Leslie Nielsen ...come on!!!) the one part I \textbf{regarded} slightly funny was the battling FBI/CIA agents, but because the audience was mainly \textbf{youngsters} they didn't understand that theme. (Positive, 59.8\%)
 \\
 \hline
Generating adversarial examples for textual entailment task \cite{Alzantot2018}&
 Premise: A runner wearing purple strives for the finish line. Hypothesis: A \textbf{runner} wants to head for the finish line. (Entailment, 86.0\%) &
 Premise: A runner wearing purple strives for the finish line. Hypothesis: A \textbf{racer} wants to head for the finish line. (Contradiction, 43.0\%)
 \\
 \hline
Back-translation \cite{Bayer2021} &
 Previously, tea had been used primarily for Buddhist monks to stay awake during mediation. &
 In the past, tea was used mostly for Buddhist monks to stay awake during the meditation.
 \\
 \hline
Contraction expansion \cite{coulombe2018} &
 I'm &
 I am
 \\
 \hline
HotFlip -- character level \cite{Ebrahimi2018} &
 South Africa's historic Soweto township marks its 100th birthday on Tuesday in a \textbf{mood} of optimism. (57\% World)&
 South Africa's historic Soweto township marks its 100th birthday on Tuesday in a \textbf{moop} of optimism. (95\% Sci/Tech)
 \\
 \hline
HotFlip -- word level \cite{Ebrahimi2018} &
 One hour photo is an \textbf{intriguing} snapshot of one man and his delusions it's just too bad it doesn't have more flashes of insight. &
 One hour photo is an \textbf{interesting} snapshot of one man and his delusions it's just too bad it doesn't have more flashes of insight
 \\
 \hline
BAE (word replacement) \cite{Garg2020} &
 The \textbf{government made} a quick decision. &
 The \textbf{(judge, doctor, captain)} made a quick decision.
 \\
 \hline
BAE (Word insertion) \cite{Garg2020} &
 The government made a quick decision. &
 The government \textbf{(officials, then, immediately)} made a quick decision. 
 \\
 \hline
AEDA \cite{Karimi2021} &
 a sad , superior human comedy played out on the back roads of life . &
 a \textbf{,} sad \textbf{.} , superior human \textbf{;} comedy \textbf{.} played \textbf{.} out on the back roads of life . 
 \\
 \hline
Character deletion \cite{Karpukhin2019} &
 wh\textbf{a}le &
 whle 
 \\
 \hline
Character insertion \cite{Karpukhin2019} &
 whale &
 w\textbf{y}hale 
 \\
 \hline
Character substitution \cite{Karpukhin2019} &
 whal\textbf{e} &
 whal\textbf{z} 
 \\
 \hline
Character swap \cite{Karpukhin2019} &
 w\textbf{ha}le &
 w\textbf{ah}le 
 \\
 \hline
DataBoost \cite{Liu2020} &
 So Cute! The baby is very lovely! &
 Look at this adorable baby! He is so cute! 
 \\
 \hline
Inversion (original premise) \cite{Min2020} &
 There are 16 El Grecos in this small collection. $\to$ This small collection contains 16 El Grecos &
 There are 16 El Grecos in this small collection. $\not\to$ 16 El Grecos contain this small collection. 
 \\
 \hline
Inversion (transformed hypothesis) \cite{Min2020} &
 There are 16 El Grecos in this small collection. → This small collection contains 16 El Grecos &
 This small collection contains 16 El Grecos. $\not\to$ 16 El Grecos contain this small collection.
 \\
 \hline
Passivization (transformed hypothesis) \cite{Min2020} &
 There are 16 El Grecos in this small collection. → This small collection contains 16 El Grecos &
 his small collection contains 16 El Grecos. $\not\to$ This small collection is contained by 16 El Grecos.
 \\
 \hline
Sentence cropping \cite{Sahin2018} &
 Her father wrote \textbf{her} a letter. &
 Her father wrote a letter.
 \\
 \hline
Sub2 \cite{Shi2021} &
 (My cat likes \textbf{milk}., I read \textbf{books}.) &
 (My cat likes \textbf{books}., I read \textbf{milk}.)
 \\
 \hline
Synonym replacement with word embeddings \cite{Wang2015} &
 Being late is terrible. &
 Be behind are bad.
 \\
 \hline
AddSentDiverse \cite{Wang2018} &
 "The Earth receives 174,000 terawatts (TW) of incoming solar radiation (insolation) at the upper atmosphere. Approximately 30\% is reflected back to space while the rest is absorbed by clouds, oceans and land masses. ... Most people around the world live in areas with insolation levels of 150 to 300 watts per square meter or 3.5 to 7.0 kWh/m2 per day." How many terawatts of solar radiation does the Earth receive? &
 "The Earth receives 174,000 terawatts (TW) of incoming solar radiation (insolation) at the upper atmosphere. Approximately 30\% is reflected back to space while the rest is absorbed by clouds, oceans and land masses. ... \textbf{The Mars receives 674000 terawatts of solar radiation.} Most people around the world live in areas with insolation levels of 150 to 300 watts per square meter or 3.5 to 7.0 kWh/m2 per day."
 \\
 \hline
EDA -- Synonym replacement \cite{Wei2019} &
 A \textbf{sad}, superior human comedy played out on the back roads of life. &
 A \textbf{lamentable}, superior human comedy played out on the backward road of life.
 \\
 \hline
EDA -- Random word insertion \cite{Wei2019} &
 A sad, superior human comedy played out on the back roads of life. &
 A sad, superior human comedy played out on \textbf{funniness} the back roads of life.
 \\
 \hline
EDA -- Random word swap \cite{Wei2019} &
 A sad, superior human comedy played out on \textbf{the} back \textbf{roads} of life. &
 A sad, superior human comedy played out on \textbf{roads} back \textbf{the} of life.
 \\
 \hline
EDA -- Random word deletion \cite{Wei2019} &
 A sad, superior human \textbf{comedy played} out on the \textbf{back} roads of life. &
 A sad, superior human out on the roads of life.
 \\
 \hline
cBERT \cite{Wu2019} &
 The actor is \textbf{{[}good{]}}. (positive) &
 The actor is \textbf{funny}.
 \\
 \hline
GPT3Mix \cite{Yoo2021} &
 It's just not very smart. (negative), It's quite an achievement to set and shoot a movie at the Cannes Film Festival and yet fail to capture its visual appeal or its atmosphere. (negative) &
 Excessively talky, occasionally absurd and much too long, Munich is a fascinating mess. (negative: 79\%)
  
\end{tabularx}
     \label{tab:DataAugmentationMethods}
     \end{center}
\end{table*}

\subsection{Supervision of data augmentation methods}
\label{Supervision of data augmentation methods}
A recent survey \cite{LiuP2020} divides data augmentation methods into supervised, semi-supervised, and unsupervised data augmentation. 

Supervised data augmentation is based on existing labeled data. Data can be transformed using techniques such as synonym replacement \cite{Wei2019}, contraction expansions \cite{coulombe2018}, or back-translation, to name a few. It is also possible to combine multiple data samples to generate one augmented data instance using methods such as mixup \cite{Zhang2017}.

Semi-supervised data augmentation combines unlabeled and labeled data. Chen et al. \cite{Chen2020} present a semi-supervised data augmentation, termed MixText, which is based on the MixMatch algorithm \cite{Berthelot2019}. MixText uses a set of unlabeled documents, which are then augmented with back-translation. Back-translation is used to produce multiple augmented documents for each unlabeled document from the original dataset. An original document and the augmented documents are fed to the model that was trained on a labeled dataset. The final label for the unlabeled original document is obtained by averaging predictions produced by the model.
Another approach to semi-supervised data augmentation is self-training. Thakur et al. \cite{Thakur2021} use a model trained on a labeled part of the dataset to label unlabeled sentences. Such augmented sentences are combined with golden annotated sentences to train the SBERT model. 

Unsupervised data augmentation uses the distribution of the data to generate new data. Bari et al. \cite{Bari2021} propose an unsupervised data augmentation framework for zero-resource cross-lingual task adaptation. The framework's goal is to adapt a task from a source language distribution to unknown target language distribution without labeled data in the target language. The authors tested the framework on cross-lingual named entity recognition (XNER) and cross-lingual natural language inference (XNLI). While this method surpasses zero-resource baselines, supervised approaches still show better performance.

\subsection{Transformation space}
\label{Transformation space}
Surveys \cite{Sahin2022} and \cite{Bayer2021} divide data augmentation methods based on how the transformations are applied on feature and data space transformations. Feature space transformations transform data in the embedding space, while data space transformations transform the raw text data.
Data space transformations can further be divided into character level, token level, phrase level, and document level transformations.

Character level augmentations have mostly been used as part of adversarial training to make the model more robust. Karpukhin et al. \cite{Karpukhin2019} experiment with character level transformations: deletion, insertion, substitution, and swap. They show that character level noise improves the performance on the test data with naturally occurring noise in the form of spelling mistakes. The output of such augmentations, if applied with small probability rates, mostly preserves the original label. The generated examples are mostly out-of-vocabulary words that are close to the original words. Such methods should be used with data representations that can model out-of-vocabulary words, such as sub-words used in BERT.

Token level augmentations perform addition, substitution, or deletion of certain tokens using different strategies. Strategies include the replacement of words with their synonyms by using resources such as WordNet \cite{Miller1995}, PPDB \cite{Pavlick2015}, or using language models to suggest substitute tokens.
Another token-level strategy is noise induction by inserting random words that do not add any label-related information \cite{Xie2020}.

Phrase-level transformations use sentence structure to add diversity to the sentence form while preserving the semantic meaning. For example, Sahin and Steedman \cite{Sahin2018} use sentence cropping to focus on subjects and objects in the sentence. Another technique the authors proposed is rotation, where parts of the sentence are moved. Authors also note that this type of augmentation is dependent on grammatical sentence structure in different languages. While some languages may permit such a change, for others the change may be invalid, merely adding noise to the model.

Document-level data augmentation tackles the problem of augmenting text consisting of multiple sentences. Yan et al. \cite{Yan2019} propose a document-level method for augmenting legal documents for a text classification task. The method randomly selects sentences from legal documents with the same label and inserts them into the current document.

While data space transformations change raw text to generate augmented instances, feature space transformations change values in the vector representation of the text. Zhang et al. \cite{Zhang2020} propose a feature space augmentation, termed SeqMix, which combines existing samples in feature space and label space to generate new samples.
The problem with feature space transformations is that they require direct access to the neural architecture because they modify the samples' embedding representation. This prevents their use in combination with a black-box model. Another issue is that in NLP, there does not necessarily exist a direct mapping between feature space and data space values. It means that values are hard to interpret because generated samples cannot be directly mapped back to text.

\subsection{Paraphrasing}
\label{Paraphrasing}
Paraphrases are sentences or phrases that convey approximately the same meaning using different wording \cite{Bhagat2013}. Data augmentation methods belonging to this category generate paraphrases of the original data which bring limited semantic change. 
The most widespread methods for generating paraphrases are replacing words with synonyms, back-translation, introducing structure changes by changing the word order, and generation of paraphrases using language models.

The simplest method is the replacement of a random word with its synonym. There are several ways to obtain synonyms for a chosen word. Typical approaches use synonym dictionaries \cite{Wei2019}, words that are close in the embedding space \cite{Wang2015}, and contextually similar words based on language models \cite{Wu2019}. Some authors select words with low TF-IDF scores and replace them with other uninformative words in the dictionary to mitigate the risk of changing the instance's class \cite{Xie2020}.

Rizos et al. \cite{Rizos2019} argue that replacing words with similar words in embedding space encourages the downstream task to place a lower emphasis on associating single words with a particular label and instead place a higher emphasis on capturing similar sequential patterns.

Back-translation or round-trip translation produces paraphrases with the help of translation models \cite{Bayer2021}. The text is first translated from the source language to the target language and then back to the language of origin. A problem with this is that the error is propagated, as there can be an error when translating from the source to the target language and again when translating back to the source language. This is especially problematic for low-resource languages with poorly performing translation models \cite{Sahin2022}. 

Language models can be used in different scenarios. One option is to use them to filter contextually unfitting synonym words based on language model output probabilities. Another possibility is to use them as the main augmentation method, where they not only substitute words with similar meaning but with words that fit the context. However, using language models for generating replacement words increases the risk for label distortion \cite{Bayer2021}. The risk can be mitigated by using label conditional language models \cite{Wu2019}. 
Some authors train label-conditioned language models from scratch, but that can be expensive and not realistic for low-data regimes. Others have proposed to reuse existing language models and combine them with reinforcement learning agent or discriminator to either condition the language model or filter out unfitting augmentations. Wu et al. \cite{Wu2019} present a pre-trained conditional language model, termed c-BERT. The authors use a pre-trained BERT model \cite{Devlin2019}, which is trained on a text classification task where segment embeddings are repurposed to encode label information.

Paraphrases can be generated by changing the structure of a sentence while preserving its label. Some of the structure-based transformations are cropping, shortening the sentence by putting the focus on the subjects and objects, or rotation, where flexible fragments of the text are moved. Shi et al. \cite{Shi2021} present structure-based augmentation, referred to as SUB2, which generates new samples by substituting sentence substructures with other substructures that have the same label. The authors test the method on part-of-speech tagging, dependency parsing, constituency parsing, and text classification tasks and achieve competitive or better performance in the few-shot setting. 

\subsection{Noising}
\label{Noising}
Noising-based data augmentation methods add noise to the original data and involve more semantic changes compared to paraphrasing methods. The main goal of such methods is to improve the robustness of the model to natural or synthetic noise. This is similar to how humans reduce the impact of noise on the understanding of the text, even if it contains some mistakes like typos or incorrect grammatical order. 
Li et al. \cite{Li2022} survey nosing methods, such as random word swap, random word deletion, random word insertion, and random word substitution. Chen et al. \cite{Chen2020} and Goyal et al. \cite{Goyal2022} survey adversarial data augmentation that can be looked at as adding noise, such as adversarial perturbation, to the original data either in data or feature space. Furthermore, Shorten et al. \cite{Shorten2021} explore feature space augmentations that include adding noise to the original data which are not included in adversarial data augmentation. 

Adversarial examples are examples that contain small changes in the input data, which are almost unrecognizable to humans, but which may mislead the algorithms into making wrong predictions. Automatic adversarial example generators can be used as data augmentation methods in order for deep learning models to be less susceptible to such alternations. On the other hand, adversarial training can disturb the true label space in the training data. For example, adversarial example generators often rely on the belief that input points that are close to each other tend to have the same label. 
The aim of the adversarial examples is to target misclassification and to make the model more robust to such changes.
Methods that use direct access to model architecture and parameters to create adversarial examples by using gradients are typically referred to as \emph{white-box methods}. 

An example of a character-level white-box method is HotFlip \cite{Ebrahimi2018}. The method estimates how much loss changes upon changing a single character. Change of a character includes the operation of substitution, insertion, and deletion. An operation that is estimated to increase the loss the most is selected.

Another example is presented by Cheng et al. \cite{Cheng2019}, who construct adversarial examples to make the neural machine translation model more robust. To construct an adversarial example, the authors propose the use of greedy search to find a substitute word that would be semantically similar to the original word in the source language but would distort the current prediction. The authors show that their approach improved translation performance and robustness on Chinese-English and English-German translation tasks.

In cases when it is not possible to directly access the model architecture and training procedure, one can resort to a black-box search for adversarial attacks. Black-box search enables adversarial example construction that is agnostic to the model architecture's internals by using grid search, random search, Bayesian optimization, and evolutionary search.

Garg and Ramakrishnan \cite{Garg2020} propose a black-box method for text classification by replacing or inserting a new token into a sentence, chosen using a pre-trained BERT model. The authors try to retain high semantic similarity with the original text by filtering the top predicted tokens by using the BERT model. The filtering is done based on a sentence similarity score computed via the Universal Sentence Encoder \cite{Cer2018}.

Wang and Bansal \cite{Wang2018} improve the robustness of a model on question-answering task by inserting a distracting sentence in the document. The sentence is distracting because it is generated as an answer to the same type of the original question but with unrelated entities.

One of the considerations when selecting adversarial data augmentation is how quickly we can construct adversarial examples, as some of the methods rely on iterative optimization, which can be a significant bottleneck for training deep learning models \cite{Shorten2021}.

Instead directly to the input, noise can be added to examples in the features pace. MODALS \cite{Cheung2021} adds Gaussian noise with zero mean and standard deviation computed across all examples in the same class. The added noise is additionally scaled with a scaling factor. 

\subsection{Sampling}
\label{Sampling}
Sampling-based data augmentation methods use the distribution of the original data to generate new data. New data can be generated by rule-based methods, label-conditioned data generation, self-training, and interpolation methods, such as mixup \cite{Li2022}.

Rule-based methods use predefined heuristics to generate new data. In contrast to paraphrasing techniques, generated data does not have to be semantically similar to original data. Min et al. \cite{Min2020} present a rule-based data augmentation method, which is used with transformer-based model and evaluated on a natural language inference (NLI) task. Their method swaps the subject and object of the original sentence and changes the verb to a passive form. The results showed that swapping subject and object reduced the model's sensitivity to word order, but passivization made a much smaller impact on the model's performance. 

Conditional data augmentation uses label information when generating new data. Two common approaches to generating conditional data are deep generative models and pre-trained language models. Liu et al. \cite{LiuP2020} find that deep generative models are less useful in practice as data augmentation is usually needed in low-data regimens and generative models require a lot of high-quality training data. 

Liu et al. \cite{Liu2020} propose a text augmentation framework, termed Data Boost, that uses an off-the-shelf GPT-2 language model \cite{Solaiman2019} to generate augmented samples. Samples are conditioned to a particular class by using reinforcement learning policy when decoding model output for the next word. This approach enables the use of a language model generator for a particular task without retraining the model from scratch. The framework enabled performance increases on Offense Detection, Sentiment Analysis, and Irony Classification tasks even when trained with just 10\% of the original datasets.   

Self-training is the approach in which a model is trained iteratively by assigning pseudo-labels to the set of unlabeled training samples with prediction greater than the defined threshold \cite{Amini2022}. As mentioned in Section \ref{Supervision of data augmentation methods}, MixText \cite{Chen2020} and SBERT \cite{Thakur2021} methods are two types of the self-training approach.

One of the earlier interpolation methods that were trying to tackle the problem of an unbalanced dataset is Synthetic Minority Over-sampling Technique (SMOTE) \cite{Chawla2002}, which works by adding interpolated samples from the minority class.

Zhang et al. \cite{Zhang2017} proposed the mixup method for data augmentation by linearly interpolating pairs of examples and their labels. Examples are interpolated as raw input vectors and labels are interpolated by using their one-hot label encoding. The method showed significant improvement in computer vision tasks by interpolating images at the pixel level. Guo and Zhang \cite{Guo2019} use mixup for sentence classification. They demonstrate how the mixup technique can be used on the word or sentence embeddings while using CNN and LSTM models. Both word and sentence-level mixup showed an increase in performance, but comparisons between the two were inconclusive as to which type yields better performance. 
Sun et al. \cite{Sun2020} incorporated the mixup technique in the transformer-based model referred to as Mixup-transformer. Unlike \cite{Zhang2017} and \cite{Guo2019}, the authors made mixup representation and activation dynamic, where they trained mixup representation and used mixup for just part of the epoch. Mixup-transformer is evaluated on the GLUE benchmark and achieves an average improvement of around 1\%. The model is also tested in the low-resource regime, where the mixup-transformer consistently outperformed BERT-large. The low-resource regime is also a more realistic scenario as practitioners will need more help from data augmentation methods when there is not enough labeled data.
Zhang et al. \cite{Zhang2020} use mixup on sequence labeling task in combination with active learning. In each iteration, their method selects unlabeled samples to annotate by using an active learning strategy. The selected samples are then augmented using their SeqMix method. The SeqMix method consists of two stages. In the first stage, eligible sequences are selected and mixed both in feature and label space. In the second stage, a discriminator is used to judge if the generated sequences are plausible or not based on their perplexity scores. Only the sequences with low perplexity are regarded as plausible ones. 
Annotated and augmented samples are added to the training set, which is used to train the final model.
Park and Caragea \cite{Park2022} go further by selecting training samples for mixup by using a strategy based on Training Dynamics. Their TDMixUp model combines easy-to-learn and ambiguous samples in the feature space. The authors test the approach on Natural Language Inference, Paraphrase Detection, and Commonsense Reasoning tasks and achieve competitive performance using a smaller subset of the training data compared with strong baselines and other data augmentation methods, such as back-translation and regular mixup.

Similar to the mixup approach, authors of the MODALS framework \cite{Cheung2021} use interpolation of samples but also extrapolation and difference transformation. Sample extrapolation is achieved by creating a new sample that lies on the line that connects the selected sample and the class centroid. Difference transformation is carried out by translating the selected sample by the difference between two random samples from the same class. The authors demonstrated improvements on multiple datasets for text, tabular, time-series, and image datasets.

Yoo et al. \cite{Yoo2021} leverage the GPT-3 \cite{Brown2020} model to mix samples from the dataset to generate new samples. The authors generate a prompt containing task description, labels, and randomly selected samples from the dataset. The output of the model are a new sample and its label.

\section{Practical challenges}
\label{Practical challenges}
Above, we described data augmentation methods. However, there are several points that one needs to keep in mind when selecting data augmentation methods and how to apply them to a language, task, and dataset at hand. This section reviews the practical challenges in applying data augmentation methods. Firstly, we will discuss at which part of the learning process to apply data augmentation (Section \ref{Offline vs online augmentation}), with which magnitude (Section \ref{Strength of data augmentation}), and how to combine multiple augmentation methods (Section \ref{Data augmentation methods stacking}). Secondly, we will discuss how data representation (Section \ref{Data representation}), task (Section \ref{Task specificity}) and language (\ref{Language specificity}) influence selection of data augmentation method. Finally, we will discuss the lack of systematic comparison of performance and computational aspects of data augmentation strategies (Section \ref{Lack of benchmarks}) and what open-source libraries for augmentation are available (Section \ref{Data augmentation libraries}). 

\subsection{Offline vs online augmentation}
\label{Offline vs online augmentation}

Data augmentation strategies can be divided into offline and online augmentation based on their role in the learning process. Data augmentation is online if embedded into the learning process so that the augmented instances are stochastically included by the learning algorithm \cite{Shorten2021}, \cite{Feng2021}. In contrast, offline data augmentation is applied to the original data independently from model training. Offline data augmentation reduces computational costs while training the model, as the augmentation is done once for all model variants. On the other hand, Shorten et al. \cite{Shorten2021} argue that data augmentation methods that are integrated into the model's learning process can better leverage the stochasticity of learning. For example, Jungiewicz and Pohl \cite{Jungiewicz2019} replace words with synonyms only if the replacement with the chosen synonym maximizes the loss of the current state of the classifier model.

Wei et al. \cite{Wei2021} explore curriculum data augmentation by gradually introducing augmented examples with original examples in training for text classification. In their setup, the authors firstly trained the model on the original data and then gradually trained it on augmented data. This approach showed improved performance compared to offline data augmentation.

The results of Kucnik and Smith \cite{Kucnik2018} show that it can be more efficient to subsample a portion of the dataset to be augmented rather than augmenting the entire dataset while keeping the performance gain as if the whole dataset was augmented. 

\subsection{Strength of data augmentation}
\label{Strength of data augmentation}
The strength of data augmentation determines the number of changes performed on an instance \cite{Shorten2021}. Wei et al. \cite{Wei2021MultiTask} define the strength of data augmentation as the ratio between the number of performed changes and the length of the sequence. The coefficient will be higher if we change two words with their synonyms instead of one and if we apply multiple transformations, such as back-translation and synonym replacement.

The strength of data augmentation needs to be optimized for a particular application as it depends on model and dataset size \cite{Cubuk2019RandAugment}. Methods that learn strength of data augmentation through training were a large part of the success of data augmentation in computer vision, as noted by Shorten et al. \cite{Shorten2021}. Successful approaches such as AutoAugmenter \cite{Cubuk2019AutoAugment}, Population-Based Augmentation \cite{Ho2019}, RandAugment \cite{Cubuk2019RandAugment} and others \cite{Yang2022} are being adapted for use in NLP but have not seen large-scale adoption \cite{Shorten2021}. Niu and Basal \cite{Niu2019} adapt AutoAugmenter for the dialogue task. The authors define a set of semantic-preserving perturbations. The perturbations are then combined by a model that learns the policy based on the reward it gets from the model trained on the augmented data and validated on the validation set.
Another example is presented by Ren et al. \cite{Ren2021}, who propose TextAutoAugment framework for learning augmentation policy by combining simpler operations for the text classification task. 

\subsection{Data augmentation methods stacking}
\label{Data augmentation methods stacking}
Research \cite{Li2022}, \cite{Bayer2021} shows that combining multiple data augmentation transformations can further improve the model's performance when compared to individual transformations. Combining multiple data augmentation methods adds multiple sources of diversity, which can improve models' generalization. As there are many augmentation methods it might be infeasible to try every combination of data augmentation methods for every NLP task. A promising direction is the automatic combination of data augmentation methods, such as TextAutoAugment \cite{Ren2021}. 

In practical applications, it might be beneficial to pair up the data augmentation method with other methods to tackle the problem of data scarcity, such as active learning. Li et al. \cite{Li2021} propose a framework that combines active learning and data augmentation on the Chinese NER task. When adding an example through active learning, the method generates additional instances by using data augmentation. This approach helps the model learn faster the decision space around the most informative samples. The paper evaluated multiple active learning strategies combined with entity replacement augmentation and token replacement augmentation while using CNN-BiLSM-CRF and Lexicon-lstm as NER models. In the best combination of model, data augmentation method, and active learning strategy the method achieved 99\% of the best deep model trained on full data using 22\% of the dataset, which is also 63\% less data compared to the active learning approach.
Zhang et al. \cite{Zhang2020} used data augmentation to multiply samples obtained by active learning strategy on the sequence labeling task. 

\subsection{Data representation}
\label{Data representation}
Data augmentation methods' performance may vary based on text representation. Some data augmentation methods, such as random character insertion, may cause the generation of out-of-vocabulary words \cite{Sahin2022}. The performance of the NLP system in that scenario would depend on how the system mitigates the issue of out-of-vocabulary words. On the other hand, recent deep learning models, such as BERT \cite{Devlin2019}, use text representations that can represent words with misspellings without having a separate word in the vocabulary.
Sahin \cite{Sahin2022}, in his comparative study on text augmentation techniques for low-resource NLP, experiments with distinct models that use various subword units: character-level\cite{Ling2015}, byte-pair encoding\cite{Sennrich2016}, and WordPiece embeddings \cite{Wu2016}. The authors show that across a number of tasks, token-level augmentation provides significant improvements for byte-pair encoding input, while character-level augmentation gives higher scores for character-level encoding and WordPiece encoding.

\subsection{Task specificity}
\label{Task specificity}
Some data augmentation methods cannot be effectively applied to every task out of the box as some transformations can remove necessary information for the task. We will explore a subset of NLP tasks and show why some of the methods are not applicable for them.

Named entity recognition task aims to identify named entities such as person, location, organization, time, etc. \cite{Yadav2018}. The identification may fail if we remove a part of a named entity by applying random word substitution or deletion on a part of the entity.

Sequence labeling task aims at assigning a label to each part of the sequence \cite{Zhiyong2020}. Zhang et al. \cite{Zhang2020} note that it is infeasible to apply context-based substitution, synonym replacement, random insertion, swap and deletion, paraphrasing, or back-translation on sequence labeling. These methods may change the labels, and determining new labels is difficult on a such granular scale without adding too much noise. On the other hand, methods such as SeqMix or models that are conditioned on the label may be able to tackle such granularity of labels without additional prior knowledge.

Similarly, text classification task is aimed at assigning a label to larger units of text such as sentences, paragraphs, or documents \cite{Bayer2021}. One example of text classification task is sentiment analysis, which aims to determine whether a document expresses positive, neutral or negative opinion. Removing random word from a document can change the meaning of the sentence and, therefore, the class. For example, removing negation "not" from the sentence "I do not like apples." would change its sentiment.

\subsection{Language specificity}
\label{Language specificity}
Data augmentation methods that work for English may not work for other languages, especially if the languages differ in their typology. In terms of morphological typology, English is an analytic language, while most low-resource languages are synthetic, as shown by \cite{Sahin2022}. Genuinely low-resource languages lack resources that are required by a broad range of data augmentation methods, while the existing resources may be of low quality, which may cause data augmentation methods to decrease performance because of added noise. Bari et al. \cite{Bari2021} note that one of the most used data augmentation methods, back-translation, may not be applicable to low-resource languages. If a good-quality translation model is not available, obtaining parallel data to train an effective machine translation model for back-translation can be more expensive for low-resource languages than annotating the target language data.

In such low-data or low-resource regimes, resource-free methods may be used. Sahin et al. \cite{Sahin2022} test character-level, token-level, and structure-based transformations on part-of-speech tagging, dependency parsing, and semantic role labeling tasks for multiple low-resource languages. Character-level transformations, such as insertion, swap, substitution, and deletion, proved to obtain the most consistent improvements. On the other hand, token-level transformations were less consistent. Synonym replacement usually lowered the performance of the model. Random word deletion and random word swap were only used on a part-of-speech tagging task, but showed positive gain on model results.
Structure-based transformations, such as cropping and rotation, were most inconsistent in improving models' results. 

Another approach to tackle low-resource regimes is to use semi-supervised and unsupervised methods discussed in Section \ref{Supervision of data augmentation methods}.

\subsection{Lack of benchmarks}
\label{Lack of benchmarks}
Automated learning of data augmentation policy for NLP is in its early stages \cite{Shorten2021}. While there is progress in automatic data augmentation without hand-picking initial transformations in computer vision \cite{Zheng2022}, most of the automated methods in NLP rely on practitioners to choose augmentation methods based on their prior knowledge of the task. When choosing augmentation methods, practitioners face a challenge as it is difficult to compare different research papers because of the lack of performance benchmarks. As Bayer et al. \cite{Bayer2021} discuss, there is a need to compare different augmentation methods in a standard way across multiple tasks based on their evaluation metrics. Currently, comparing the methods is difficult and requires adaptation as many authors omit the information on how they applied the augmentation, sampled examples from the dataset, and how many examples their method creates per each original example.

While performance benchmarks can provide insights into which data augmentation method works better, they fall short of capturing the computational aspect of applying data augmentation. Computing is cheaper than human annotation, but there are applications such as crisis intervention for which one needs to train a classifier in a very limited time and where time-consuming data augmentation methods may not be applicable \cite{Bayer2021}.

\subsection{Data augmentation libraries}
\label{Data augmentation libraries}
Recent advances in data augmentation for natural language processing have led to the development of a number of open-source libraries. Using such libraries can help accelerate the usage of different data augmentation methods in product development.

AugLy library \cite{Papakipos2022}, OpenAttack toolkit \cite{Zeng2021}, and TextAttack framework \cite{Morris2020} are focused on data augmentations based on adversarial attacks for making the models more robust. While TextAttack and OpenAttack are focused on text data, AugLy supports other data modalities such as audio, image, and video.
NLPaug \cite{Ma2019} offers simple-to-use transformations for text and audio data. NL-Augmenter \cite{Dhole2021} provides a wide range of text transformations with a focus on task specificity. As discussed in Section \ref{Task specificity}, library authors argue that some transformations may be irrelevant or degrade performance if they are not task-specific. 
TextFlint \cite{Wang2021} is a library for multilingual robustness evaluation, which also offers a wide range of text transformations. Similarly to libraries based on adversarial attacks, Authors of NL-Augmenter and TextFlint libraries suggest using data augmentation methods on validation and test sets to determine the robustness of the model.

\section{Conclusion}
Data augmentation is a set of strategies for generating new data based on available data. This paper presented an overview of data augmentation methods and their practical challenges.
Data augmentation proved to be a low-cost solution to the data scarcity problem and a way to make the model more robust.
While effective in many use cases, it is still not easy to determine which data augmentation will be successful and how it should be applied to the particular use case. 
Future directions might include an analysis of how to apply data augmentation more effectively by selecting appropriate data instances and stacking data augmentation methods.

\bibliographystyle{IEEEtran}
\bibliography{IEEEabrv, main}

\begin{thebibliography}{10}
\providecommand{\url}[1]{#1}
\csname url@samestyle\endcsname
\providecommand{\newblock}{\relax}
\providecommand{\bibinfo}[2]{#2}
\providecommand{\BIBentrySTDinterwordspacing}{\spaceskip=0pt\relax}
\providecommand{\BIBentryALTinterwordstretchfactor}{4}
\providecommand{\BIBentryALTinterwordspacing}{\spaceskip=\fontdimen2\font plus
\BIBentryALTinterwordstretchfactor\fontdimen3\font minus
  \fontdimen4\font\relax}
\providecommand{\BIBforeignlanguage}[2]{{%
\expandafter\ifx\csname l@#1\endcsname\relax
\typeout{** WARNING: IEEEtran.bst: No hyphenation pattern has been}%
\typeout{** loaded for the language `#1'. Using the pattern for}%
\typeout{** the default language instead.}%
\else
\language=\csname l@#1\endcsname
\fi
#2}}
\providecommand{\BIBdecl}{\relax}
\BIBdecl

\bibitem{Tan2020}
Z.~Tan, S.~Wang \emph{et~al.}, ``Neural machine translation: A review of
  methods, resources, and tools,'' \emph{AI Open}, vol.~1, pp. 5--21, 1 2020.

\bibitem{Devlin2019}
\BIBentryALTinterwordspacing
J.~Devlin, M.-W. Chang \emph{et~al.}, ``{BERT}: Pre-training of deep
  bidirectional transformers for language understanding,'' in \emph{Proceedings
  of the 2019 Conference of the North {A}merican Chapter of the Association for
  Computational Linguistics: Human Language Technologies, Volume 1 (Long and
  Short Papers)}.\hskip 1em plus 0.5em minus 0.4em\relax Minneapolis,
  Minnesota: Association for Computational Linguistics, Jun. 2019, pp.
  4171--4186. [Online]. Available: \url{https://aclanthology.org/N19-1423}
\BIBentrySTDinterwordspacing

\bibitem{Minaee2021}
\BIBentryALTinterwordspacing
S.~Minaee, N.~Kalchbrenner \emph{et~al.}, ``Deep learning--based text
  classification,'' \emph{ACM Computing Surveys (CSUR)}, vol.~54, 4 2021.
  [Online]. Available: \url{https://dl.acm.org/doi/10.1145/3439726}
\BIBentrySTDinterwordspacing

\bibitem{Sahin2022}
\BIBentryALTinterwordspacing
G.~G. Şahin, ``To augment or not to augment? a comparative study on text
  augmentation techniques for low-resource {NLP},'' \emph{Computational
  Linguistics}, vol.~48, pp. 5--42, 4 2022. [Online]. Available:
  \url{https://aclanthology.org/2022.cl-1.2}
\BIBentrySTDinterwordspacing

\bibitem{Feng2021}
\BIBentryALTinterwordspacing
S.~Y. Feng, V.~Gangal \emph{et~al.}, ``A survey of data augmentation approaches
  for {NLP},'' \emph{Findings of the Association for Computational Linguistics:
  ACL-IJCNLP 2021}, pp. 968--988, 2021. [Online]. Available:
  \url{https://aclanthology.org/2021.findings-acl.84}
\BIBentrySTDinterwordspacing

\bibitem{Li2022}
\BIBentryALTinterwordspacing
B.~Li, Y.~Hou, and W.~Che, ``Data augmentation approaches in natural language
  processing: A survey,'' \emph{AI Open}, 2022. [Online]. Available:
  \url{https://www.sciencedirect.com/science/article/pii/S2666651022000080}
\BIBentrySTDinterwordspacing

\bibitem{Bayer2021}
\BIBentryALTinterwordspacing
B.~Markus, K.~Marc-André, and R.~Christian, ``A survey on data augmentation
  for text classification,'' \emph{ACM Computing Surveys}, 7 2021. [Online].
  Available: \url{https://dl.acm.org/doi/10.1145/3544558}
\BIBentrySTDinterwordspacing

\bibitem{Hedderich2021}
\BIBentryALTinterwordspacing
M.~A. Hedderich, L.~Lange \emph{et~al.}, ``A survey on recent approaches for
  natural language processing in low-resource scenarios,'' in \emph{Proceedings
  of the 2021 Conference of the North American Chapter of the Association for
  Computational Linguistics: Human Language Technologies}.\hskip 1em plus 0.5em
  minus 0.4em\relax Online: Association for Computational Linguistics, Jun.
  2021, pp. 2545--2568. [Online]. Available:
  \url{https://aclanthology.org/2021.naacl-main.201}
\BIBentrySTDinterwordspacing

\bibitem{Shorten2019}
\BIBentryALTinterwordspacing
C.~Shorten and T.~M. Khoshgoftaar, ``A survey on image data augmentation for
  deep learning,'' \emph{Journal of Big Data}, vol.~6, pp. 1--48, 12 2019.
  [Online]. Available:
  \url{https://journalofbigdata.springeropen.com/articles/10.1186/s40537-019-0197-0}
\BIBentrySTDinterwordspacing

\bibitem{Garcia2018}
\BIBentryALTinterwordspacing
A.~Hernández-García and P.~König, ``Data augmentation instead of explicit
  regularization,'' \emph{arXiv:1806.03852}, 6 2018. [Online]. Available:
  \url{https://arxiv.org/abs/1806.03852v5}
\BIBentrySTDinterwordspacing

\bibitem{Cheung2021}
\BIBentryALTinterwordspacing
T.-H. Cheung and D.-Y. Yeung, ``{\{}MODALS{\}}: Modality-agnostic automated
  data augmentation in the latent space,'' in \emph{International Conference on
  Learning Representations}, 2021. [Online]. Available:
  \url{https://openreview.net/forum?id=XjYgR6gbCEc}
\BIBentrySTDinterwordspacing

\bibitem{Shorten2021}
\BIBentryALTinterwordspacing
C.~Shorten, T.~M. Khoshgoftaar, and B.~Furht, ``Text data augmentation for deep
  learning,'' \emph{Journal of Big Data 2021 8:1}, vol.~8, pp. 1--34, 7 2021.
  [Online]. Available:
  \url{https://journalofbigdata.springeropen.com/articles/10.1186/s40537-021-00492-0}
\BIBentrySTDinterwordspacing

\bibitem{Sharma2020}
\BIBentryALTinterwordspacing
S.~Sharma, Y.~Zhang \emph{et~al.}, ``Data augmentation for discrimination
  prevention and bias disambiguation,'' in \emph{Proceedings of the AAAI/ACM
  Conference on AI, Ethics, and Society}, ser. AIES '20.\hskip 1em plus 0.5em
  minus 0.4em\relax New York, NY, USA: Association for Computing Machinery,
  2020, p. 358–364. [Online]. Available:
  \url{https://doi.org/10.1145/3375627.3375865}
\BIBentrySTDinterwordspacing

\bibitem{Park2018}
\BIBentryALTinterwordspacing
J.~H. Park, J.~Shin, and P.~Fung, ``Reducing gender bias in abusive language
  detection,'' in \emph{Proceedings of the 2018 Conference on Empirical Methods
  in Natural Language Processing}.\hskip 1em plus 0.5em minus 0.4em\relax
  Brussels, Belgium: Association for Computational Linguistics, Oct.-Nov. 2018,
  pp. 2799--2804. [Online]. Available: \url{https://aclanthology.org/D18-1302}
\BIBentrySTDinterwordspacing

\bibitem{Wei2019}
\BIBentryALTinterwordspacing
J.~Wei and K.~Zou, ``Eda: Easy data augmentation techniques for boosting
  performance on text classification tasks,'' \emph{EMNLP-IJCNLP 2019 - 2019
  Conference on Empirical Methods in Natural Language Processing and 9th
  International Joint Conference on Natural Language Processing, Proceedings of
  the Conference}, pp. 6382--6388, 2019. [Online]. Available:
  \url{https://aclanthology.org/D19-1670}
\BIBentrySTDinterwordspacing

\bibitem{Karimi2021}
\BIBentryALTinterwordspacing
A.~Karimi, L.~Rossi, and A.~Prati, ``Aeda: An easier data augmentation
  technique for text classification,'' \emph{Findings of the Association for
  Computational Linguistics, Findings of ACL: EMNLP 2021}, pp. 2748--2754,
  2021. [Online]. Available:
  \url{https://aclanthology.org/2021.findings-emnlp.234}
\BIBentrySTDinterwordspacing

\bibitem{Longpre2020}
\BIBentryALTinterwordspacing
S.~Longpre, Y.~Wang, and C.~DuBois, ``How effective is task-agnostic data
  augmentation for pretrained transformers?'' \emph{Findings of the Association
  for Computational Linguistics Findings of ACL: EMNLP 2020}, pp. 4401--4411,
  2020. [Online]. Available:
  \url{https://aclanthology.org/2020.findings-emnlp.394}
\BIBentrySTDinterwordspacing

\bibitem{Simard2003}
P.~Y. Simard, D.~Steinkraus, and J.~C. Platt, ``Best practices for
  convolutional neural networks applied to visual document analysis,''
  \emph{Proceedings of the International Conference on Document Analysis and
  Recognition, ICDAR}, vol. 2003-January, pp. 958--963, 2003.

\bibitem{Ko2015}
T.~Ko, V.~Peddinti \emph{et~al.}, ``Audio augmentation for speech
  recognition,'' \emph{Proceedings of the Annual Conference of the
  International Speech Communication Association, INTERSPEECH}, vol.
  2015-January, pp. 3586--3589, 2015.

\bibitem{LiuP2020}
P.~Liu, X.~Wang \emph{et~al.}, ``A survey of text data augmentation,''
  \emph{Proceedings - 2020 International Conference on Computer Communication
  and Network Security, CCNS 2020}, pp. 191--195, 8 2020.

\bibitem{Alzantot2018}
\BIBentryALTinterwordspacing
M.~Alzantot, Y.~Sharma \emph{et~al.}, ``Generating natural language adversarial
  examples,'' \emph{Proceedings of the 2018 Conference on Empirical Methods in
  Natural Language Processing, EMNLP 2018}, pp. 2890--2896, 2018. [Online].
  Available: \url{https://aclanthology.org/D18-1316}
\BIBentrySTDinterwordspacing

\bibitem{Wu2019}
\BIBentryALTinterwordspacing
X.~Wu, S.~Lv \emph{et~al.}, ``Conditional bert contextual augmentation,''
  \emph{Lecture Notes in Computer Science (including subseries Lecture Notes in
  Artificial Intelligence and Lecture Notes in Bioinformatics)}, vol. 11539
  LNCS, pp. 84--95, 2019. [Online]. Available:
  \url{https://link.springer.com/chapter/10.1007/978-3-030-22747-0\_7}
\BIBentrySTDinterwordspacing

\bibitem{Liu2020}
\BIBentryALTinterwordspacing
R.~Liu, G.~Xu \emph{et~al.}, ``Data boost: Text data augmentation through
  reinforcement learning guided conditional generation,'' \emph{EMNLP 2020 -
  2020 Conference on Empirical Methods in Natural Language Processing,
  Proceedings of the Conference}, pp. 9031--9041, 2020. [Online]. Available:
  \url{https://aclanthology.org/2020.emnlp-main.726}
\BIBentrySTDinterwordspacing

\bibitem{coulombe2018}
C.~Coulombe, ``Text data augmentation made simple by leveraging nlp cloud
  apis,'' \emph{arXiv:1812.04718}, 2018.

\bibitem{Ebrahimi2018}
\BIBentryALTinterwordspacing
J.~Ebrahimi, A.~Rao \emph{et~al.}, ``{H}ot{F}lip: White-box adversarial
  examples for text classification,'' in \emph{Proceedings of the 56th Annual
  Meeting of the Association for Computational Linguistics (Volume 2: Short
  Papers)}.\hskip 1em plus 0.5em minus 0.4em\relax Melbourne, Australia:
  Association for Computational Linguistics, Jul. 2018, pp. 31--36. [Online].
  Available: \url{https://aclanthology.org/P18-2006}
\BIBentrySTDinterwordspacing

\bibitem{Garg2020}
\BIBentryALTinterwordspacing
S.~Garg and G.~Ramakrishnan, ``{BAE}: {BERT}-based adversarial examples for
  text classification,'' in \emph{Proceedings of the 2020 Conference on
  Empirical Methods in Natural Language Processing (EMNLP)}.\hskip 1em plus
  0.5em minus 0.4em\relax Online: Association for Computational Linguistics,
  Nov. 2020, pp. 6174--6181. [Online]. Available:
  \url{https://aclanthology.org/2020.emnlp-main.498}
\BIBentrySTDinterwordspacing

\bibitem{Karpukhin2019}
\BIBentryALTinterwordspacing
V.~Karpukhin, O.~Levy \emph{et~al.}, ``Training on synthetic noise improves
  robustness to natural noise in machine translation,'' in \emph{Proceedings of
  the 5th Workshop on Noisy User-generated Text (W-NUT 2019)}.\hskip 1em plus
  0.5em minus 0.4em\relax Hong Kong, China: Association for Computational
  Linguistics, Nov. 2019, pp. 42--47. [Online]. Available:
  \url{https://aclanthology.org/D19-5506}
\BIBentrySTDinterwordspacing

\bibitem{Min2020}
\BIBentryALTinterwordspacing
J.~Min, R.~T. McCoy \emph{et~al.}, ``Syntactic data augmentation increases
  robustness to inference heuristics,'' in \emph{Proceedings of the 58th Annual
  Meeting of the Association for Computational Linguistics}.\hskip 1em plus
  0.5em minus 0.4em\relax Online: Association for Computational Linguistics,
  Jul. 2020, pp. 2339--2352. [Online]. Available:
  \url{https://aclanthology.org/2020.acl-main.212}
\BIBentrySTDinterwordspacing

\bibitem{Sahin2018}
\BIBentryALTinterwordspacing
G.~G. {\c{S}}ahin and M.~Steedman, ``Data augmentation via dependency tree
  morphing for low-resource languages,'' in \emph{Proceedings of the 2018
  Conference on Empirical Methods in Natural Language Processing}.\hskip 1em
  plus 0.5em minus 0.4em\relax Brussels, Belgium: Association for Computational
  Linguistics, Oct.-Nov. 2018, pp. 5004--5009. [Online]. Available:
  \url{https://aclanthology.org/D18-1545}
\BIBentrySTDinterwordspacing

\bibitem{Shi2021}
\BIBentryALTinterwordspacing
H.~Shi, K.~Livescu, and K.~Gimpel, ``Substructure substitution: Structured data
  augmentation for {NLP},'' in \emph{Findings of the Association for
  Computational Linguistics: ACL-IJCNLP 2021}.\hskip 1em plus 0.5em minus
  0.4em\relax Online: Association for Computational Linguistics, Aug. 2021, pp.
  3494--3508. [Online]. Available:
  \url{https://aclanthology.org/2021.findings-acl.307}
\BIBentrySTDinterwordspacing

\bibitem{Wang2015}
\BIBentryALTinterwordspacing
W.~Y. Wang and D.~Yang, ``That{'}s so annoying!!!: A lexical and frame-semantic
  embedding based data augmentation approach to automatic categorization of
  annoying behaviors using {\#}petpeeve tweets,'' in \emph{Proceedings of the
  2015 Conference on Empirical Methods in Natural Language Processing}.\hskip
  1em plus 0.5em minus 0.4em\relax Lisbon, Portugal: Association for
  Computational Linguistics, Sep. 2015, pp. 2557--2563. [Online]. Available:
  \url{https://aclanthology.org/D15-1306}
\BIBentrySTDinterwordspacing

\bibitem{Wang2018}
\BIBentryALTinterwordspacing
Y.~Wang and M.~Bansal, ``Robust machine comprehension models via adversarial
  training,'' \emph{CoRR}, vol. abs/1804.06473, 2018. [Online]. Available:
  \url{http://arxiv.org/abs/1804.06473}
\BIBentrySTDinterwordspacing

\bibitem{Yoo2021}
\BIBentryALTinterwordspacing
K.~M. Yoo, D.~Park \emph{et~al.}, ``{GPT}3{M}ix: Leveraging large-scale
  language models for text augmentation,'' in \emph{Findings of the Association
  for Computational Linguistics: EMNLP 2021}.\hskip 1em plus 0.5em minus
  0.4em\relax Punta Cana, Dominican Republic: Association for Computational
  Linguistics, Nov. 2021, pp. 2225--2239. [Online]. Available:
  \url{https://aclanthology.org/2021.findings-emnlp.192}
\BIBentrySTDinterwordspacing

\bibitem{Zhang2017}
\BIBentryALTinterwordspacing
H.~Zhang, M.~Cisse \emph{et~al.}, ``mixup: Beyond empirical risk
  minimization,'' \emph{6th International Conference on Learning
  Representations, ICLR 2018 - Conference Track Proceedings}, 10 2017.
  [Online]. Available: \url{https://arxiv.org/abs/1710.09412v2}
\BIBentrySTDinterwordspacing

\bibitem{Chen2020}
\BIBentryALTinterwordspacing
J.~Chen, Z.~Yang, and D.~Yang, ``{M}ix{T}ext: Linguistically-informed
  interpolation of hidden space for semi-supervised text classification,'' in
  \emph{Proceedings of the 58th Annual Meeting of the Association for
  Computational Linguistics}.\hskip 1em plus 0.5em minus 0.4em\relax Online:
  Association for Computational Linguistics, Jul. 2020, pp. 2147--2157.
  [Online]. Available: \url{https://aclanthology.org/2020.acl-main.194}
\BIBentrySTDinterwordspacing

\bibitem{Berthelot2019}
D.~Berthelot, N.~Carlini \emph{et~al.}, \emph{MixMatch: A Holistic Approach to
  Semi-Supervised Learning}.\hskip 1em plus 0.5em minus 0.4em\relax Red Hook,
  NY, USA: Curran Associates Inc., 2019.

\bibitem{Thakur2021}
\BIBentryALTinterwordspacing
N.~Thakur, N.~Reimers \emph{et~al.}, ``Augmented {SBERT}: Data augmentation
  method for improving bi-encoders for pairwise sentence scoring tasks,'' in
  \emph{Proceedings of the 2021 Conference of the North American Chapter of the
  Association for Computational Linguistics: Human Language
  Technologies}.\hskip 1em plus 0.5em minus 0.4em\relax Online: Association for
  Computational Linguistics, Jun. 2021, pp. 296--310. [Online]. Available:
  \url{https://aclanthology.org/2021.naacl-main.28}
\BIBentrySTDinterwordspacing

\bibitem{Bari2021}
\BIBentryALTinterwordspacing
M.~S. Bari, T.~Mohiuddin, and S.~Joty, ``{UXLA}: A robust unsupervised data
  augmentation framework for zero-resource cross-lingual {NLP},'' in
  \emph{Proceedings of the 59th Annual Meeting of the Association for
  Computational Linguistics and the 11th International Joint Conference on
  Natural Language Processing (Volume 1: Long Papers)}.\hskip 1em plus 0.5em
  minus 0.4em\relax Online: Association for Computational Linguistics, Aug.
  2021, pp. 1978--1992. [Online]. Available:
  \url{https://aclanthology.org/2021.acl-long.154}
\BIBentrySTDinterwordspacing

\bibitem{Miller1995}
\BIBentryALTinterwordspacing
G.~A. Miller, ``Wordnet: A lexical database for english,'' \emph{Commun. ACM},
  vol.~38, no.~11, p. 39–41, nov 1995. [Online]. Available:
  \url{https://doi.org/10.1145/219717.219748}
\BIBentrySTDinterwordspacing

\bibitem{Pavlick2015}
\BIBentryALTinterwordspacing
E.~Pavlick, P.~Rastogi \emph{et~al.}, ``{PPDB} 2.0: Better paraphrase ranking,
  fine-grained entailment relations, word embeddings, and style
  classification,'' in \emph{Proceedings of the 53rd Annual Meeting of the
  Association for Computational Linguistics and the 7th International Joint
  Conference on Natural Language Processing (Volume 2: Short Papers)}.\hskip
  1em plus 0.5em minus 0.4em\relax Beijing, China: Association for
  Computational Linguistics, Jul. 2015, pp. 425--430. [Online]. Available:
  \url{https://aclanthology.org/P15-2070}
\BIBentrySTDinterwordspacing

\bibitem{Xie2020}
Q.~Xie, Z.~Dai \emph{et~al.}, ``Unsupervised data augmentation for consistency
  training,'' in \emph{Proceedings of the 34th International Conference on
  Neural Information Processing Systems}, ser. NIPS'20.\hskip 1em plus 0.5em
  minus 0.4em\relax Red Hook, NY, USA: Curran Associates Inc., 2020.

\bibitem{Yan2019}
G.~Yan, Y.~Li \emph{et~al.}, ``Data augmentation for deep learning of judgment
  documents,'' in \emph{Intelligence Science and Big Data Engineering. Big Data
  and Machine Learning}, Z.~Cui, J.~Pan \emph{et~al.}, Eds.\hskip 1em plus
  0.5em minus 0.4em\relax Cham: Springer International Publishing, 2019, pp.
  232--242.

\bibitem{Zhang2020}
\BIBentryALTinterwordspacing
R.~Zhang, Y.~Yu, and C.~Zhang, ``{S}eq{M}ix: Augmenting active sequence
  labeling via sequence mixup,'' in \emph{Proceedings of the 2020 Conference on
  Empirical Methods in Natural Language Processing (EMNLP)}.\hskip 1em plus
  0.5em minus 0.4em\relax Online: Association for Computational Linguistics,
  Nov. 2020, pp. 8566--8579. [Online]. Available:
  \url{https://aclanthology.org/2020.emnlp-main.691}
\BIBentrySTDinterwordspacing

\bibitem{Bhagat2013}
\BIBentryALTinterwordspacing
R.~Bhagat and E.~Hovy, ``{What Is a Paraphrase?}'' \emph{Computational
  Linguistics}, vol.~39, no.~3, pp. 463--472, 09 2013. [Online]. Available:
  \url{https://doi.org/10.1162/COLI\_a\_00166}
\BIBentrySTDinterwordspacing

\bibitem{Rizos2019}
\BIBentryALTinterwordspacing
G.~Rizos, K.~Hemker, and B.~Schuller, ``Augment to prevent: Short-text data
  augmentation in deep learning for hate-speech classification,'' in
  \emph{Proceedings of the 28th ACM International Conference on Information and
  Knowledge Management}, ser. CIKM '19.\hskip 1em plus 0.5em minus 0.4em\relax
  New York, NY, USA: Association for Computing Machinery, 2019, p. 991–1000.
  [Online]. Available: \url{https://doi.org/10.1145/3357384.3358040}
\BIBentrySTDinterwordspacing

\bibitem{Goyal2022}
S.~Goyal, S.~Doddapaneni \emph{et~al.}, ``A survey in adversarial defences and
  robustness in {NLP},'' \emph{ArXiv}, vol. abs/2203.06414, 2022.

\bibitem{Cheng2019}
\BIBentryALTinterwordspacing
Y.~Cheng, L.~Jiang, and W.~Macherey, ``Robust neural machine translation with
  doubly adversarial inputs,'' in \emph{Proceedings of the 57th Annual Meeting
  of the Association for Computational Linguistics}.\hskip 1em plus 0.5em minus
  0.4em\relax Florence, Italy: Association for Computational Linguistics, Jul.
  2019, pp. 4324--4333. [Online]. Available:
  \url{https://aclanthology.org/P19-1425}
\BIBentrySTDinterwordspacing

\bibitem{Cer2018}
\BIBentryALTinterwordspacing
D.~Cer, Y.~Yang \emph{et~al.}, ``Universal sentence encoder,'' \emph{CoRR},
  vol. abs/1803.11175, 2018. [Online]. Available:
  \url{http://arxiv.org/abs/1803.11175}
\BIBentrySTDinterwordspacing

\bibitem{Solaiman2019}
\BIBentryALTinterwordspacing
I.~Solaiman, M.~Brundage \emph{et~al.}, ``Release strategies and the social
  impacts of language models,'' \emph{CoRR}, vol. abs/1908.09203, 2019.
  [Online]. Available: \url{http://arxiv.org/abs/1908.09203}
\BIBentrySTDinterwordspacing

\bibitem{Amini2022}
M.-R. Amini, V.~Feofanov \emph{et~al.}, ``Self-training: A survey,''
  \emph{ArXiv}, vol. abs/2202.12040, 2022.

\bibitem{Chawla2002}
N.~V. Chawla, K.~W. Bowyer \emph{et~al.}, ``Smote: Synthetic minority
  over-sampling technique,'' \emph{J. Artif. Int. Res.}, vol.~16, no.~1, p.
  321–357, jun 2002.

\bibitem{Guo2019}
\BIBentryALTinterwordspacing
H.~Guo, Y.~Mao, and R.~Zhang, ``Augmenting data with mixup for sentence
  classification: An empirical study,'' \emph{ArXiv}, 5 2019. [Online].
  Available: \url{https://arxiv.org/abs/1905.08941v1}
\BIBentrySTDinterwordspacing

\bibitem{Sun2020}
\BIBentryALTinterwordspacing
L.~Sun, C.~Xia \emph{et~al.}, ``Mixup-transformer: Dynamic data augmentation
  for {NLP} tasks,'' in \emph{Proceedings of the 28th International Conference
  on Computational Linguistics}.\hskip 1em plus 0.5em minus 0.4em\relax
  Barcelona, Spain (Online): International Committee on Computational
  Linguistics, Dec. 2020, pp. 3436--3440. [Online]. Available:
  \url{https://aclanthology.org/2020.coling-main.305}
\BIBentrySTDinterwordspacing

\bibitem{Park2022}
\BIBentryALTinterwordspacing
S.~Y. Park and C.~Caragea, ``A data cartography based {M}ix{U}p for pre-trained
  language models,'' in \emph{Proceedings of the 2022 Conference of the North
  American Chapter of the Association for Computational Linguistics: Human
  Language Technologies}.\hskip 1em plus 0.5em minus 0.4em\relax Seattle,
  United States: Association for Computational Linguistics, Jul. 2022, pp.
  4244--4250. [Online]. Available:
  \url{https://aclanthology.org/2022.naacl-main.314}
\BIBentrySTDinterwordspacing

\bibitem{Brown2020}
\BIBentryALTinterwordspacing
T.~B. Brown, B.~Mann \emph{et~al.}, ``Language models are few-shot learners,''
  \emph{CoRR}, vol. abs/2005.14165, 2020. [Online]. Available:
  \url{https://arxiv.org/abs/2005.14165}
\BIBentrySTDinterwordspacing

\bibitem{Jungiewicz2019}
\BIBentryALTinterwordspacing
M.~Jungiewicz and A.~Smywinski-Pohl, ``Towards textual data augmentation for
  neural networks: synonyms and maximum loss,'' \emph{Computer Science},
  vol.~20, no.~1, Mar. 2019. [Online]. Available:
  \url{https://journals.agh.edu.pl/csci/article/view/3023}
\BIBentrySTDinterwordspacing

\bibitem{Wei2021}
\BIBentryALTinterwordspacing
J.~Wei, C.~Huang \emph{et~al.}, ``Few-shot text classification with triplet
  networks, data augmentation, and curriculum learning,'' in \emph{Proceedings
  of the 2021 Conference of the North American Chapter of the Association for
  Computational Linguistics: Human Language Technologies}.\hskip 1em plus 0.5em
  minus 0.4em\relax Online: Association for Computational Linguistics, Jun.
  2021, pp. 5493--5500. [Online]. Available:
  \url{https://aclanthology.org/2021.naacl-main.434}
\BIBentrySTDinterwordspacing

\bibitem{Kucnik2018}
\BIBentryALTinterwordspacing
M.~Kuchnik and V.~Smith, ``Efficient augmentation via data subsampling,''
  \emph{CoRR}, vol. abs/1810.05222, 2018. [Online]. Available:
  \url{http://arxiv.org/abs/1810.05222}
\BIBentrySTDinterwordspacing

\bibitem{Wei2021MultiTask}
\BIBentryALTinterwordspacing
J.~Wei, C.~Huang \emph{et~al.}, ``Text augmentation in a multi-task view,'' in
  \emph{Proceedings of the 16th Conference of the European Chapter of the
  Association for Computational Linguistics: Main Volume}.\hskip 1em plus 0.5em
  minus 0.4em\relax Online: Association for Computational Linguistics, Apr.
  2021, pp. 2888--2894. [Online]. Available:
  \url{https://aclanthology.org/2021.eacl-main.252}
\BIBentrySTDinterwordspacing

\bibitem{Cubuk2019RandAugment}
\BIBentryALTinterwordspacing
E.~D. Cubuk, B.~Zoph \emph{et~al.}, ``Randaugment: Practical data augmentation
  with no separate search,'' \emph{CoRR}, vol. abs/1909.13719, 2019. [Online].
  Available: \url{http://arxiv.org/abs/1909.13719}
\BIBentrySTDinterwordspacing

\bibitem{Cubuk2019AutoAugment}
\BIBentryALTinterwordspacing
------, ``Autoaugment: Learning augmentation policies from data,'' in
  \emph{Proceedings of the IEEE/CVF Conference on Computer Vision and Pattern
  Recognition (CVPR)}, 2019. [Online]. Available:
  \url{https://arxiv.org/pdf/1805.09501.pdf}
\BIBentrySTDinterwordspacing

\bibitem{Ho2019}
\BIBentryALTinterwordspacing
D.~Ho, E.~Liang \emph{et~al.}, ``Population based augmentation: Efficient
  learning of augmentation policy schedules,'' \emph{CoRR}, vol.
  abs/1905.05393, 2019. [Online]. Available:
  \url{http://arxiv.org/abs/1905.05393}
\BIBentrySTDinterwordspacing

\bibitem{Yang2022}
Z.~Yang, R.~O. Sinnott \emph{et~al.}, ``A survey of automated data augmentation
  algorithms for deep learning-based image classication tasks,'' \emph{arXiv
  preprint arXiv:2206.06544}, 2022.

\bibitem{Niu2019}
\BIBentryALTinterwordspacing
T.~Niu and M.~Bansal, ``Automatically learning data augmentation policies for
  dialogue tasks,'' in \emph{Proceedings of the 2019 Conference on Empirical
  Methods in Natural Language Processing and the 9th International Joint
  Conference on Natural Language Processing (EMNLP-IJCNLP)}.\hskip 1em plus
  0.5em minus 0.4em\relax Hong Kong, China: Association for Computational
  Linguistics, Nov. 2019, pp. 1317--1323. [Online]. Available:
  \url{https://aclanthology.org/D19-1132}
\BIBentrySTDinterwordspacing

\bibitem{Ren2021}
\BIBentryALTinterwordspacing
S.~Ren, J.~Zhang \emph{et~al.}, ``Text {A}uto{A}ugment: Learning compositional
  augmentation policy for text classification,'' in \emph{Proceedings of the
  2021 Conference on Empirical Methods in Natural Language Processing}.\hskip
  1em plus 0.5em minus 0.4em\relax Online and Punta Cana, Dominican Republic:
  Association for Computational Linguistics, Nov. 2021, pp. 9029--9043.
  [Online]. Available: \url{https://aclanthology.org/2021.emnlp-main.711}
\BIBentrySTDinterwordspacing

\bibitem{Li2021}
\BIBentryALTinterwordspacing
Q.~Li, Z.~Huang \emph{et~al.}, ``A framework of data augmentation while active
  learning for chinese named entity recognition,'' in \emph{Knowledge Science,
  Engineering and Management: 14th International Conference, KSEM 2021, Tokyo,
  Japan, August 14–16, 2021, Proceedings, Part II}.\hskip 1em plus 0.5em
  minus 0.4em\relax Berlin, Heidelberg: Springer-Verlag, 2021, p. 88–100.
  [Online]. Available: \url{https://doi.org/10.1007/978-3-030-82147-0\_8}
\BIBentrySTDinterwordspacing

\bibitem{Ling2015}
\BIBentryALTinterwordspacing
W.~Ling, C.~Dyer \emph{et~al.}, ``Finding function in form: Compositional
  character models for open vocabulary word representation,'' in
  \emph{Proceedings of the 2015 Conference on Empirical Methods in Natural
  Language Processing}.\hskip 1em plus 0.5em minus 0.4em\relax Lisbon,
  Portugal: Association for Computational Linguistics, Sep. 2015, pp.
  1520--1530. [Online]. Available: \url{https://aclanthology.org/D15-1176}
\BIBentrySTDinterwordspacing

\bibitem{Sennrich2016}
\BIBentryALTinterwordspacing
R.~Sennrich, B.~Haddow, and A.~Birch, ``Neural machine translation of rare
  words with subword units,'' in \emph{Proceedings of the 54th Annual Meeting
  of the Association for Computational Linguistics (Volume 1: Long
  Papers)}.\hskip 1em plus 0.5em minus 0.4em\relax Berlin, Germany: Association
  for Computational Linguistics, Aug. 2016, pp. 1715--1725. [Online].
  Available: \url{https://aclanthology.org/P16-1162}
\BIBentrySTDinterwordspacing

\bibitem{Wu2016}
Y.~Wu, M.~Schuster \emph{et~al.}, ``Google's neural machine translation system:
  Bridging the gap between human and machine translation,'' \emph{ArXiv}, vol.
  abs/1609.08144, 2016.

\bibitem{Yadav2018}
\BIBentryALTinterwordspacing
V.~Yadav and S.~Bethard, ``A survey on recent advances in named entity
  recognition from deep learning models,'' in \emph{Proceedings of the 27th
  International Conference on Computational Linguistics}.\hskip 1em plus 0.5em
  minus 0.4em\relax Santa Fe, New Mexico, USA: Association for Computational
  Linguistics, Aug. 2018, pp. 2145--2158. [Online]. Available:
  \url{https://aclanthology.org/C18-1182}
\BIBentrySTDinterwordspacing

\bibitem{Zhiyong2020}
\BIBentryALTinterwordspacing
Z.~He, Z.~Wang \emph{et~al.}, ``A survey on recent advances in sequence
  labeling from deep learning models,'' \emph{CoRR}, vol. abs/2011.06727, 2020.
  [Online]. Available: \url{https://arxiv.org/abs/2011.06727}
\BIBentrySTDinterwordspacing

\bibitem{Zheng2022}
\BIBentryALTinterwordspacing
Y.~Zheng, Z.~Zhang \emph{et~al.}, ``Deep autoaugment,'' in \emph{International
  Conference on Learning Representations}, 2022. [Online]. Available:
  \url{https://openreview.net/forum?id=St-53J9ZARf}
\BIBentrySTDinterwordspacing

\bibitem{Papakipos2022}
\BIBentryALTinterwordspacing
Z.~Papakipos, J.~Bitton, and M.~Ai, ``Augly: Data augmentations for
  robustness,'' \emph{arXiv:2201.06494v1}, 1 2022. [Online]. Available:
  \url{https://arxiv.org/abs/2201.06494v1}
\BIBentrySTDinterwordspacing

\bibitem{Zeng2021}
\BIBentryALTinterwordspacing
G.~Zeng, F.~Qi \emph{et~al.}, ``{O}pen{A}ttack: An open-source textual
  adversarial attack toolkit,'' in \emph{Proceedings of the 59th Annual Meeting
  of the Association for Computational Linguistics and the 11th International
  Joint Conference on Natural Language Processing: System
  Demonstrations}.\hskip 1em plus 0.5em minus 0.4em\relax Online: Association
  for Computational Linguistics, Aug. 2021, pp. 363--371. [Online]. Available:
  \url{https://aclanthology.org/2021.acl-demo.43}
\BIBentrySTDinterwordspacing

\bibitem{Morris2020}
\BIBentryALTinterwordspacing
J.~X. Morris, E.~Lifland \emph{et~al.}, ``Textattack: A framework for
  adversarial attacks, data augmentation, and adversarial training in {NLP},''
  \emph{arXiv:2005.05909v4}, pp. 119--126, 4 2020. [Online]. Available:
  \url{https://arxiv.org/abs/2005.05909v4}
\BIBentrySTDinterwordspacing

\bibitem{Ma2019}
E.~Ma, ``{NLP} augmentation,'' https://github.com/makcedward/nlpaug, 2019.

\bibitem{Dhole2021}
\BIBentryALTinterwordspacing
K.~D. Dhole, V.~Gangal \emph{et~al.}, ``Nl-augmenter: A framework for
  task-sensitive natural language augmentation,'' \emph{Nivranshu Pasricha},
  vol.~20, p.~75, 12 2021. [Online]. Available:
  \url{https://arxiv.org/abs/2112.02721v1}
\BIBentrySTDinterwordspacing

\bibitem{Wang2021}
\BIBentryALTinterwordspacing
X.~Wang, Q.~Liu \emph{et~al.}, ``{T}ext{F}lint: Unified multilingual robustness
  evaluation toolkit for natural language processing,'' in \emph{Proceedings of
  the 59th Annual Meeting of the Association for Computational Linguistics and
  the 11th International Joint Conference on Natural Language Processing:
  System Demonstrations}.\hskip 1em plus 0.5em minus 0.4em\relax Online:
  Association for Computational Linguistics, Aug. 2021, pp. 347--355. [Online].
  Available: \url{https://aclanthology.org/2021.acl-demo.41}
\BIBentrySTDinterwordspacing

\end{thebibliography}
\end{document}